\documentclass[a4paper]{article}
\usepackage{graphicx}
\usepackage{twocolceurws}
\usepackage{hyperref}
\usepackage{amsmath}

\title{SemEval-2017 Task 4: Sentiment Analysis in Twitter using BERT}

\author{
Rupak Kumar Das \\ Department of Computer Science\\
University of Minnesota Duluth\\
                Duluth, MN 55812 \\ das00057@d.umn.edu
\and
Dr. Ted Pedersen \\ Department of Computer Science\\
University of Minnesota Duluth\\
                Duluth, MN 55812 \\ tpederse@d.umn.edu }

\institution{}

\begin{document}
\maketitle

\begin{abstract}
This paper uses the BERT model, which is a transformer-based architecture, to solve task 4A, English Language, Sentiment Analysis in Twitter of SemEval2017. BERT is a very powerful large language model for classification tasks when the amount of training data is small. For this experiment, we have used the BERT{\textsubscript{\tiny BASE}} model, which has 12 hidden layers. This model provides better accuracy, precision, recall, and f1 score than the Naive Bayes baseline model. It performs better in binary classification subtasks than the multi-class classification subtasks. We also considered all kinds of ethical issues during this experiment, as Twitter data contains personal and sensible information. The dataset and code used in our experiment can be found in this \href{https://github.umn.edu/DAS00057/Research-Project-UMD-CS-5642}{github} repository.
\end{abstract}

\section{Introduction}
\label{introduction}

Sentiment Analysis is used to detect whether a piece of text expresses a positive or a negative or a neutral opinion. The piece of text can be a product review, a blog post, an event, etc. Applications of sentiment analysis are vast. Sentiment analysis is very useful in social media monitoring as the use of social media like Facebook, Twitter, Skype, and WhatsApp is increasing every day. This analysis can help us to gain an overview of public opinion. The study of public opinion is now important, and the rise of social media content has created new opportunities for us. Social media, especially Twitter, is very popular for research because it contains a variety of topics and public opinion. By sentiment analysis of customer opinion, sellers not only can see what the customers think of them but also can see what customers think about the competitors. There are 2 challenges in Twitter sentiment analysis:

\begin{itemize}
  \item The limit of the length of the message on Twitter is 140 characters.
  \item Use of informal language.
\end{itemize}

The Twitter Sentiment analysis has been running at SemEval since 2013. SemEval is an international workshop on Semantic Evaluation. Previously, it was known as SensEval. This workshop contains an ongoing series of computational sentiment analysis evaluations. Since 2013, more than 40 teams have participated in this Sentiment Analysis on the Twitter workshop. Previous editions of the SemEval mainly involved Polarity (Positive or Negative) classification \cite{rosenthal2017semeval}. The workshop of 2014 was a rerun of 2013, but they introduced three new test sets: 1) regular tweets, 2) sarcastic tweets, and 3) LiveJournal sentences \cite{velichkov2014fmi}. In SemEval-2015 task 10, three new sub-tasks asked to predict 1) the sentiment towards a topic in a single tweet, 2) the overall sentiment towards a topic in a set of tweets, 3) the degree of prior polarity of a phrase \cite{rosenthal2019semeval2015} are added. As usual, SemEval-2016 Task 4 was a rerun of SemEval-2015 Task 10 with two important changes: 1) Replacing classification with quantification, 2) Replacing a two-point scale (Positive / Negative) or three-point scale (Positive + Neutral + Negative) problem with five-point scale (Very Positive, Positive, Neutral, Negative, Very Negative) \cite{nakov2019semeval}.

In this paper, we perform the re-run of the sub-tasks in the SemEval-2016 Task. Although there are 5 sub-tasks, we will work on Sub-tasks A, B, and C. Besides, it contains polarity classification in another language and User Demographic information (e.g., age, location) to analyze the impact on improving sentiment analysis. Here, Arabic is introduced as the new language. But we work on only the English language for those sub-tasks.

\section{Classification AND Quantification}

SemEval-2017 Task 4 contains tweet quantification tasks as well as tweet classification tasks on 2-point and 5-point scales. Sentiment classification is a task of detecting whether a text or opinion (e.g., a product review, a blog post, an editorial, etc.) is expressing a positive, negative, or neutral opinion. This type of classification technique has many applications in market research, political science, social science, and others \cite{gao2016classification}. But it is not that important to find whether a specific person has a positive or negative view of that topic. Instead, it is more important to find the distribution of positive and negative tweets about a given topic. This is known as quantification in data mining and related fields. Quantification is not a byproduct of classification. A good classifier is not necessarily a good quantifier, and vice versa \cite{forman2008quantifying}.

\section{Task Description}

SemEval-2017 Task 4 consists of five subtasks for both English and Arabic languages.

\begin{itemize}
  \item \textbf{Subtask A:} Given a message, classify whether the message is of positive, negative, or neutral sentiment.
  
  \item \textbf{Subtask B:} Given a message and a topic, classify based on a two-point scale (positive or negative).
  
  \item \textbf{Subtask C:} Given a message and a topic, classify based on a five-point scale.
  
  \item \textbf{Subtask D:} Given a set of tweets about a given topic, estimate the distribution of the tweets across a two-point scale (Positive or Negative classes)
  
  \item \textbf{Subtask E:} Given a set of tweets about a given topic, estimate the distribution of the tweets across a five-point scale.
\end{itemize}

We will work on sub-tasks A, B, and C for only the English language.

\section{Related Works}
In \cite{cliche2017bb_twtr}, the authors used Convolutional Neural Network and Long Short-Term Memory (LSTMs) networks for the Twitter sentiment classifier. They used 100 million unlabeled tweets to pre-trained the word embeddings, then used them in CNN and LSTM. Word2vec, FastText, and GloVe are used as unsupervised learning algorithms. The embeddings learned in this phase were not good enough. They fine-tuned the embeddings via a distant training phase. The final training stage uses human-labeled data provided by SemEval-2017. To reduce variance and boost accuracy, they grouped 10 CNNs and 10 LSTMs together. The Final result shows that FastText and Word2vec give a higher score than the GloVe unsupervised algorithm.

\cite{jabreel2017sitaka} describes a system named SiTAKA, which is used in English and Arabic languages, Sentiment Analysis in Twitter of SemEval 2017 subtask 4A. The authors used word2vec and SSWEu pre-trained embedding models in this system. After some standard pre-processing methods (Normalization, Tokenization, Negation), their system used five types of features: basic text, lexicon, syntactic, cluster, and Word embeddings. Support Vector Machine (SVM) is used for training. Their system ranked 8th among 38 systems in the English language tweets and ranked 2nd among 8 systems in the Arabic language tweets.

In \cite{gonzalez2017elirf}, the authors used Convolutional and Recurrent Neural Network (CRNN) for both English and Arabic languages. Their system combines three Convolutional and Recurrent Neural Networks (CRNN). In-domain embeddings (word2vec), out-domain embedding (word2vec), and sequences of the polarity of the words (sequence of C-dimensional one-hot vectors) are used as the input of this system. The output of those three CRNNs is then concatenated and used as the input of a fully connected Multilayer Perceptron (MLP). In subtask A, the accuracy achieved the 24th position. In subtask B, their position was lower because no information on the topic was included in the model. Subtask C achieved 7th position, and in subtask D, the result was very poor. But in subtask E, they ranked 4th.

\section{Proposed Methodology}

We used the BERT model from the hugging face framework, which is available in \href{https://huggingface.co/transformers/model_doc/bert.html}{Hugging-Face Website}. We used BERT{\textsubscript{\tiny BASE}}  as BERT{\textsubscript{\tiny LARGE}} is more complicated and requires more computational resources.

\subsection{BERT Model}
A Bidirectional Encoder Representation from the Transformers model (BERT) is first proposed in \cite{devlin2018bert}. There are two steps in this framework: Pre-training and Fine-tuning. This model is trained with unlabeled data over different pre-training tasks. It can be fine-tuned using task-specific labeled data.

The BERT model is actually a multi-layer bidirectional Transformer encoder. The Transformer architecture is described in \cite{vaswani2017attention}. It is an encoder-decoder network that uses self-attention on the encoder side and attention on the decoder side. The Transformer reads entire sequences of tokens at once, while LSTMs read sequentially. There are two types of BERT models available for use. The first one is BERT{\textsubscript{\tiny BASE}, which contains 12 layers, and the other one is BERT{\textsubscript{\tiny LARGE}}, which has 24 layers in the encoder stack. In the Transformer, the number of layers in an encoder is 6. After the encoder layer, both BERT models have a feedforward network with 768 and 1024 hidden layers, respectively. Those two models have more self-attention heads (12 and 16 respectively) than Transformer. BERT{\textsubscript{BASE}} contains 110M parameters while BERT{\textsubscript{LARGE}} contains 340M parameters. Both BERT models are shown in figure \ref{figure1}.

        \begin{figure}[ht]
            \centering
            \includegraphics[width=\columnwidth]{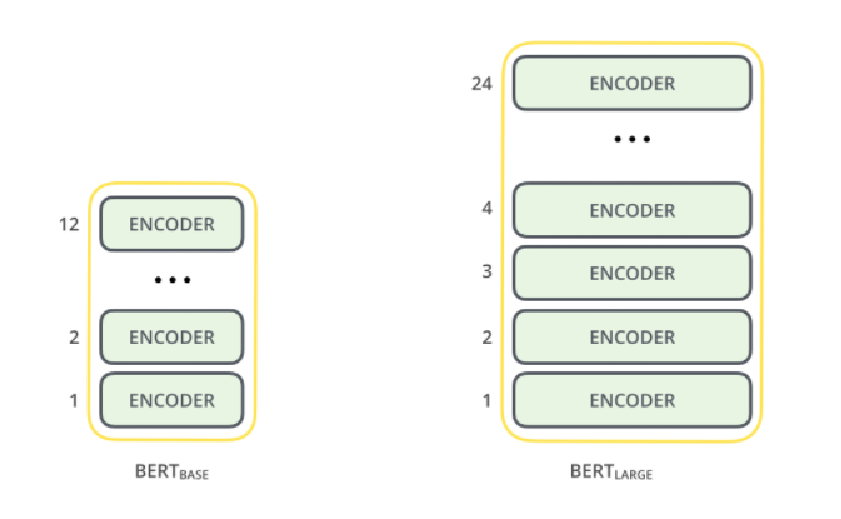}
            \caption{Base and Large BERT Encoder \cite{bertpicJayAlammar}}
            \label{figure1}
        \end{figure}

This model has 30,000 token vocabularies. It takes ([CLS]) token as input first, then it is followed by a sequence of words as input. Here ([CLS]) is a classification token. It then passes the input to the above layers. Each layer applies self-attention and passes the result through a feedforward network. After that, it hands off to the next encoder. The model outputs a vector of hidden size (768 for BERT{\textsubscript{\tiny BASE}} and 1024 for BERT{\textsubscript{\tiny LARGE}} ). If we want to output a classifier from this model, we can take the output corresponding to the [CLS] token.

BERT is pre-trained using Masked Language Modeling (MLM) and Next Sentence Prediction (NSP) on a large corpus from Wikipedia.  BERT was trained by masking 15\text{\%} of the tokens with the goal of guessing those words. For the pre-training corpus, authors used BooksCorpus, which contains 800M words, and English Wikipedia, which contains 2,500M words.

Fine-tuning a BERT model is very important because it is relatively inexpensive compared to pre-training. It takes very little data to fine-tune and provides better accuracy than other deep-learning models like CNN and LSTM. In this research project, we took the pre-trained BERT model, added an untrained layer of neurons on the end, and then trained our new model for this Twitter sentiment classification task. The pre-train and fine-tuning process is shown in figure \ref{figure2}

        \begin{figure}[ht]
            \centering
            \includegraphics[width=\columnwidth]{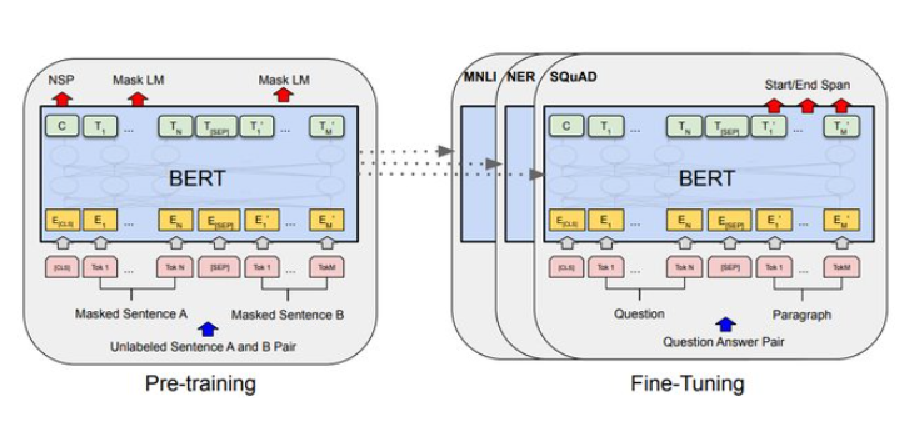}
            \caption{Pre-train and Fine-tuning procedure for BERT \cite{devlin2018bert}}
            \label{figure2}
        \end{figure}

BERT is helpful for different Natural Language Processing tasks like classification tasks, Name Entity Recognition, Part of Speech tagging, Question Answering, etc. However, it is not useful for Language Models, Language Translation, or Text Generation. The BERT model is large and takes time for fine-tuning and inferencing.

\subsection{Baseline Method}

After implementing a model, we want to know whether it performs better than the other models, especially if there is a simpler or more tractable approach. This approach is referred to as the baseline. For a baseline proof-of-concept model, we used a Naive Bayes Classifier, which is often a good choice as a baseline model. This classifier is a simple classifier for the classification based on probabilities of a particular event. It is normally used in text classification problems. It takes less training time and less training data, less CPU and memory consumption.

\subsection{DataSets and Evaluation Measures}

For this experiment, we used DataSets provided by SemEval 2017. A summary of the datasets is given in Table 1. For this experiment, Colab Notebook with Tesla T4 (16 GB) GPU was used. 

\begin{table}[hbt]
    \begin{tabular}{ |p{2cm}||p{1.2cm}|p{1.2cm}| p{1.2cm}| }
     \hline
    \multicolumn{4}{|c|}{Dataset} \\
    \hline
    Task& Train data&Test data& DevTest data\\
    \hline
    SubTask-A   & 5868    &20632& 2000\\
    SubTask-B&   4309  & 10551& 1417\\
    SubTask-C &5868 & 20632& 2000\\
    \hline
    \end{tabular}
    \label{dataset}
    \caption{SemEval 2017 dataset for Twitter Sentiment Analysis Task}
\end{table}

For the evaluation measures for the three subtasks, we used accuracy, precision, recall, and F1 score.

Accuracy is defined as the ratio of the number of correct predictions to the total number of predictions.

\[Accuracy = \frac{TP+TF}{TP+TF+FP+FN}\]

The precision is the ratio of the number of true positives to the number of false positives.

\[Precision = \frac{TP}{(TP+NP)}\]

The recall is the ratio of the number of true positives to the number of false negatives.

\[Recall = \frac{TP}{TP+FN}\]

The formula for the standard F1-score is the harmonic mean of the precision and recall. A perfect model has an F-score of 1.

\[F1 = \frac{2*Precision*Recall}{Precision+Recall}\]

\begin{table*}[hbt]
\centering
    \begin{tabular}{ |p{3cm}||p{2cm}|p{2cm}| p{2cm}| p{2cm}|}
    \hline
    Model  &Accuracy &Precision    &Recall &F1 Score\\
    \hline
    BERT-SubTask-A   & 0.6337  &0.6296  &0.6202 &0.5931\\
    \hline
    BERT-SubTask-B  &0.8969  & 0.8381   &0.8882 &0.8484\\
    \hline
    BERT-SubTask-C &0.5422 & 0.9004 &0.5422 &0.6346\\
    \hline
    Baseline A  &0.3686 &0.3176 &0.3490 &0.2131\\
    \hline
    Baseline B  &0.7783 &0.7783 &0.8900 &0.8753\\
    \hline
    Baseline C  &0.3836 &0.2153 &0.2016 &0.1142\\
    \hline
    \end{tabular}
    \label{results}
    \caption{SemEval 2017 results of three subtasks with BERT model along with baseline method}
\end{table*}

\section{Result and Discussion}
Accuracy, precision, recall, and f1 score are used to find the overall performance of the proposed BERT model. The summary of all three tasks is given in Table 2. For every subtask, the BERT model performed much better than the baseline (Naive Bayes) method. It performed better for binary classification than multi-class classification. Because of the imbalanced data and small training dataset, the accuracy of the baseline method is poor. The baseline method also performed better for binary classification rather than multi-class classification. As BERT uses a pre-trained model for fine-tuning, its performance is much better than the baseline.

\section{Ethical Consideration}

Social media activity has been increasing rapidly in the world for the last few years. There are 2.5 billion users of Facebook, Google+, and Twitter \cite{williams2017towards}. Those accounts produce a lot of personal data daily. Twitter is the most widely used social media platform in the world, and researchers collect data from it because it is a relatively easy task. It is an open platform, so maximum posts are publicly available, and researchers can collect a huge amount of data in a short period of time. It provides several application programming interfaces (APIs) that allow real-time access to vast amounts of content. Data extraction from Twitter API contains personal information. The publication of user’s Twitter data in research can raise significant questions for research ethics. Publishing tweets can cause new or larger forms of attention, and they might be at risk from their participation in research if their identities are revealed. It is better to take consent from the users for data collection. However, researchers normally work with large datasets, in which it is not possible to obtain informed consent from all the users.

Ethical consideration should always take priority over academic findings. It is suggested in \cite{webb2017ethical} that qualitative analysis should have composite or paraphrased data rather than publish actual tweets. It is also possible to develop a tool that creates tweets automatically while creating a Twitter-based sentiment analysis tool. In \cite{freitas2016empirical}, the authors have presented reverse engineering socialbot strategies in Twitter. A socialbot is a software program that simulates human behavior automatically on a social media platform. For this experiment, the authors had to create 120 social bots, which created a few thousand social links on Twitter.

So, in this research project, we consider all ethical issues strictly. All kinds of research integrity and data privacy are maintained here. Only qualitative analysis is considered for publishing this research result.

\section{Conclusion and Future Work}
In this paper, we used the BERT model, which is a transformer-based architecture for the sentiment analysis of the messages posted on Twitter. A pre-trained BERT model was fine-tuned with the data provided by SemEval-2017 for this task. There were a total of 5 subtasks under Task 4 in 2 different languages. We worked on subtasks A, B, and C in the English language. In the future, we will also work on Subtasks D and E in English and another language.

\bibliographystyle{unsrt} 
\bibliography{paper2}





\end{document}